\documentclass[runningheads]{llncs}

\usepackage{eccv}

\usepackage{eccvabbrv}

\usepackage{graphicx}
\graphicspath{{./Images/}}
\usepackage{booktabs}
\usepackage{amsmath}
\usepackage{tikz}
\usetikzlibrary{arrows}
\usetikzlibrary{graphs}
\usetikzlibrary{positioning}
\usetikzlibrary{quotes}
\usetikzlibrary{fit}
\usepackage{mathtools}

\usepackage{pgfplots}
\usepackage{pgfplotstable}
\pgfplotsset{compat=1.9}

\usepackage{svg}
\usepackage{cellspace}

\usepackage[accsupp]{axessibility}  % Improves PDF readability for those with disabilities.

\usepackage[normalem]{ulem}

\usepackage{hyperref}

\usepackage{orcidlink}

\newcommand{\SMT}{\mathit{SMT}}
\newcommand{\RFW}{\mathit{RFW}}
\newcommand{\CW}{\mathit{CW}}
\newcommand{\MA}{\mathit{MA}}

\begin{document}

\title{Visual Motif Identification:\\Elaboration of a Curated Comparative Dataset and Classification Methods} % Maybe talk about using CLIP or other powerful tools in title

\titlerunning{Visual Motif Identification}

\author{Adam Phillips \orcidlink{0009-0007-3329-389X} \and
Daniel Grandes Rodriguez \orcidlink{0009-0001-3767-2471} \and
Miriam Sánchez-Manzano \orcidlink{0000-0002-4103-9118} \and
Alan Salvadó \orcidlink{0000-0001-8282-2021} \and
Manuel Garin \orcidlink{0000-0001-9596-8257} \and
Gloria Haro \orcidlink{0000-0002-8194-8092} \and
Coloma Ballester \orcidlink{0000-0001-6535-7367}
}

\authorrunning{A. Phillips et al.}

\institute{Universitat Pompeu Fabra, Barcelona, Spain}

\maketitle

\begin{abstract}
  In cinema, visual motifs are recurrent iconographic compositions that carry artistic or aesthetic significance. Their use throughout the history of visual arts and media is interesting to researchers and filmmakers alike. Our goal in this work is to recognise and classify these motifs by proposing a new machine learning model that uses a custom dataset to that end. We show how features extracted from a CLIP model can be leveraged by using a shallow network and an appropriate loss to classify images into 20 different motifs, with surprisingly good results: an $F_1$-score of 0.91 on our test set. We also present several ablation studies justifying the input features, architecture and hyperparameters used.
\end{abstract}

\section{Introduction}
\label{sec:intro}

In cinema, artists often make visual references to previous films or works of art, or compose frames and scenes following codes that permeate throughout the history of the medium. For example, the \textit{Pietà} is a recurring theme in many forms of art, iconographically represented by Mary holding the dead body of Christ, such as in the \textit{Madonna della Pietà} (1499) by Michelangelo. This pose - the horizontality of the deceased body, piously held in arms - has since been used in many other works, including in many films or media news, to represent the lamentation of someone's suffering. Being able to automatically detect the presence of such visual motifs would help alleviate workload for researchers in the fields of iconology and visual culture, and could also incite artists and content creators to integrate specific motifs while being aware of their use throughout art history.

\begin{figure}[tb]
  \centering
  \includegraphics[width=0.8\textwidth]{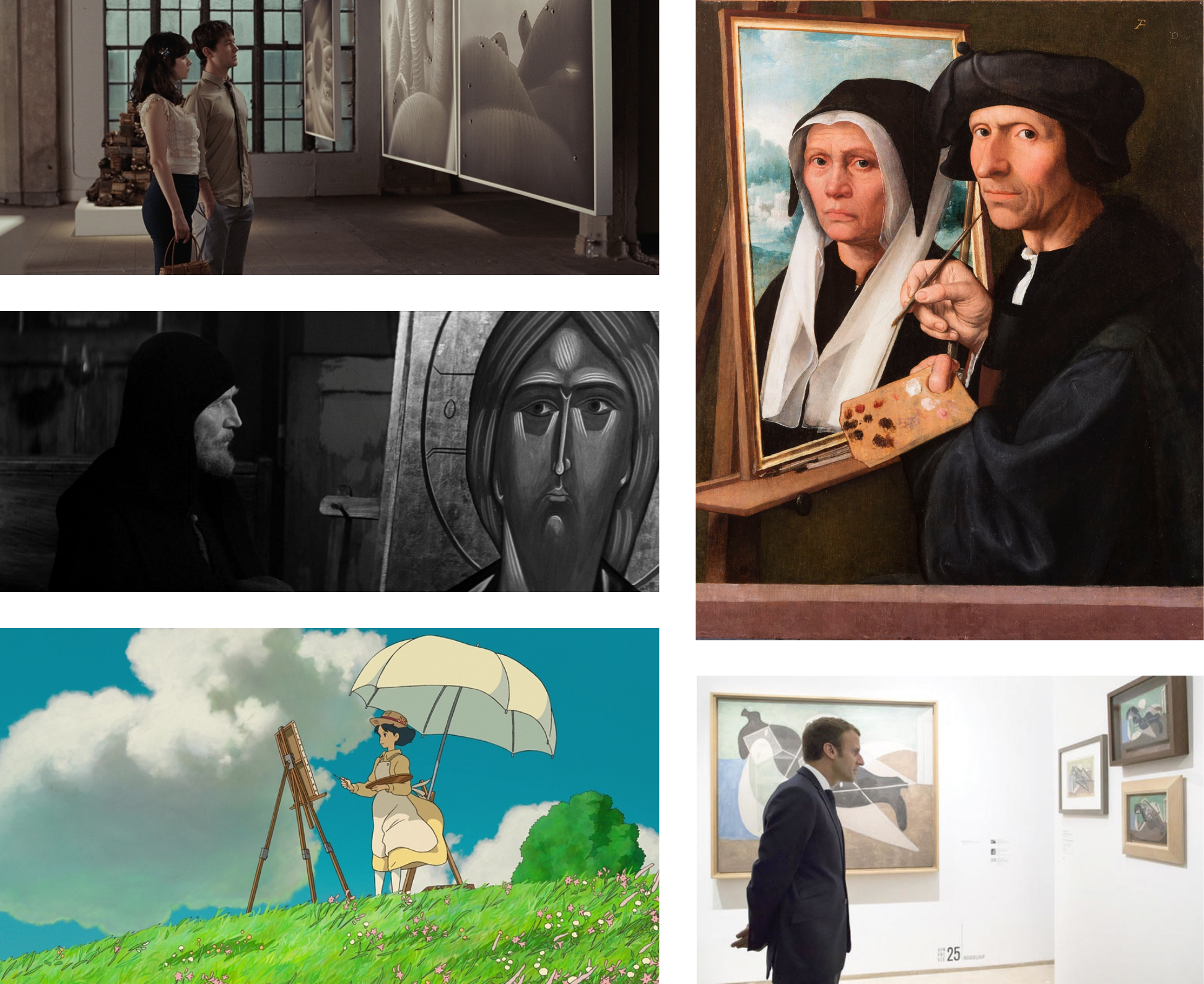}
  \caption{A selection of images from our dataset representing the motif \textit{Facing the Image}. On the left, three stills from films of different styles and periods: from top to bottom, \textit{(500) Days of Summer} (2009) by Marc Webb, \textit{Andrei Rublev} (1966) by Andrei Tarkovsky, and \textit{The Wind Rises} (2013) by Hayao Miyazaki. On the right, the painting \textit{Jacob Cornelisz. van Oostsanen Painting a Portrait of His Wife} (circa 1530-1550) by Dirck Jacobsz and a press photography of French President Emmanuel Macron at the Picasso National Museum in Paris.}
  \label{fig:facingtheimage}
\end{figure}

With this goal in mind, as a team containing art historians, we first determined a select subset of visual motifs to focus our automatic identification on, and have been developing a corresponding benchmark dataset of images, which we refer to as the \emph{Curated Comparative Dataset}. This dataset was made with the objective of including and representing the characteristics, variations and subtleties of these motifs, as completely as possible, by showcasing examples from all types of media, period, origin, \etc, as illustrated in \Cref{fig:facingtheimage}.

We then used this dataset to train and test a classification network, with the objective of determining the visual motifs (of our subset) that are present in any given image. Due to the relatively small size of our image dataset, our method involves extracting its representative features using powerful, pre-existing foundation models such as CLIP (Contrastive Language-Image Pre-training), and training our own network with an appropriate loss using these features as input. In the following, we present our method and results, and show that a surprisingly simple classification head is enough to efficiently determine the motifs of the images of the testing set, based only on their CLIP features. We compare results of models trained on several other types of features, and present the possible extensions and improvements of our model which will be the focus of our future work.

\section{Related Work}

\subsubsection{Visual Motifs} Despite the use of computer vision tools being a growing interest in the domain of digital humanities \cite{smits2023CLIPandDH, wevers2021scenedetectiondeboer, oiva2024newsreelanalysis, bhargav2019intertitles}, the automatic detection of visual motifs is, to the best of our knowledge, a new topic of research. Visual motifs, however, are a key element to the iconographic analysis and understanding of visual media, and have been studied not only applied to cinema \cite{ballo2000motifs, ballobergala2016motifs}, but also in adjacent art forms like photography \cite{berger2016motifs, sontag1977motifs}, TV series \cite{ballo2018motivostvseries}, or the public sphere \cite{salvado2020powermotifs, garin2021economy}. The prevalence of motifs in visual culture, which have been shown to have similarities across different types of media \cite{salvado2023motifs}, shows that they are an essential building block of how works of art were and are created, and that being able to identify and analyse them automatically would greatly help researchers and creators alike.

\subsubsection{Image Understanding}

Although the identification of visual motifs specifically is a new topic in computer vision, it is closely related to several other fields, such as image understanding. CLIP \cite{radford2021CLIP} (Contrastive Language-Image Pre-training), as well as the open source versions by OpenCLIP \cite{cherti2023openclip, ilharco2021openclipsoftware, Radford2021openclip, sun2023evaclip}, are models that have been trained on large image/text datasets to learn visual information from natural language supervision. Thus, the features extracted from images with their visual models have been shown to be powerful, and adaptable to many a task, such as image captioning \cite{mokady2021imagecaptioning}, image classification \cite{abdelfattah2023imageclassification}, and image generation \cite{ramesh2022texttoimage}. The same can be said of Vision Transformers trained on large amounts of images, such as DINOv2 \cite{oquab2024dinov2} which has shown to give state-of-the-art features for many benchmarks. Closely related to visual motif detection is that of object detection, since many motifs involve specific objects. Tools such as Detectron2 \cite{wu2019detectron2}, with state-of-the-art object detection capabilities, could therefore be adapted to help with the task of identifying visual motifs. Other endeavours go even further and aim to fully describe a scene's elements and composition, through scene graphs and specified datasets \cite{krishna2017visualgenome, ma2023composition}.

\subsubsection{AI and Art}

Adapting tools and models developed with images to be used in works of art, particularly in paintings, is an active area of research \cite{castellano2021patternpaintingsoverview}. Indeed, the state-of-the-art tools in computer vision tend to be trained for photographs, which follow very different conventions than paintings, making the task much harder. Methods include adapting pre-existing tools, as in \cite{gonthier2018objectdetectioninpaintings, elgammal2018analyseartstyle, castellano2021visuallinkretrieval, madhu2020compositionalstructures}, including using CLIP \cite{posthumus2022clipart}, and developing new purpose-built datasets, as in \cite{shen2019patternrepetitionpaintings, reshetnikov2022deart}.

\begin{figure}[tb]
\centering
\begin{tikzpicture}
  \centering
  \begin{axis}[
        ybar, axis on top,
        width = \textwidth,
        height = 4cm,
        bar width = 0.25cm,
        ymajorgrids, tick align=inside,
        major grid style={draw=black},
        enlarge y limits={value=.1,upper},
        ymin=0, ymax=1000,
        axis x line*=bottom,
        axis y line*=right,
        y axis line style={opacity=0},
        tickwidth=0pt,
        enlarge x limits=true,
        %legend style={
        %    at={(0.5,-0.2)},
        %    anchor=north,
        %    legend columns=-1,
        %    /tikz/every even column/.append style={column sep=0.5cm}
        %},
        %ylabel={Number of images},
        symbolic x coords={Autograph, Brawl, Duel, Facing the Image, Family Photo, Handshake, Horizon, Hug, Kiss, Leader Walks Alone, Line, Maze, Mirror, Pep Talk, Pietà, Presidential Box, Shadow, Signing Contract, Stairs, Woman in the Window},
       xtick=data,
       nodes near coords={
        \pgfmathprintnumber[precision=0]{\pgfplotspointmeta}
       },
       x tick label style={rotate=45, anchor = north east},
       tick label style={font=\scriptsize}
    ]
    \addplot [draw=none, fill=red!30, font = \scriptsize] coordinates {
      (Autograph, 727)
      (Brawl, 632)
      (Duel, 721)
      (Facing the Image, 336)
      (Family Photo, 578)
      (Handshake, 471)
      (Horizon, 480)
      (Hug, 601)
      (Kiss, 642)
      (Leader Walks Alone, 461)
      (Line, 688)
      (Maze, 387)
      (Mirror, 456)
      (Pep Talk, 705)
      (Pietà, 469)
      (Presidential Box, 683)
      (Shadow, 359)
      (Signing Contract, 729)
      (Stairs, 379)
      (Woman in the Window, 381)};
    %\legend{First Fix}
  \end{axis}
  \end{tikzpicture}
\caption{Graph showing the 20 motifs of our proposed \textit{Curated Comparative Dataset}, and the corresponding number of image samples per motif (proportional bar heights)}
\label{fig:motifcount}
\end{figure}
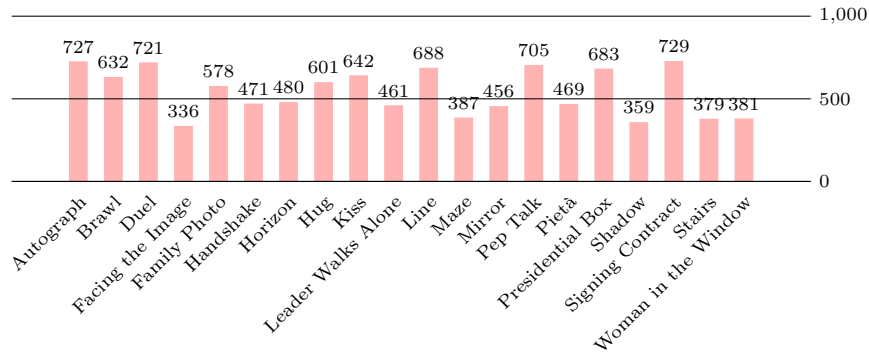 

\section{Dataset} 
\label{sec:dataset}

The current version of our \textit{Curated Comparative Dataset}, which we are planning to eventually make publicly available, is composed of 10,760 images, representing a total of 20 motifs. In particular, we have focused on the following visual motifs: \textit{Autograph}, \textit{Brawl}, \textit{Duel}, \textit{Facing the Image}, \textit{Family Photo}, \textit{Handshake}, \textit{Horizon}, \textit{Hug}, \textit{Kiss}, \textit{Leader Walks Alone}, \textit{Line}, \textit{Maze}, \textit{Mirror}, \textit{Pep Talk}, \textit{Pietà}, \textit{Presidential Box}, \textit{Shadow}, \textit{Signing Contract}, \textit{Stairs}, and \textit{Woman in the Window}. The number of images per motif is presented in \Cref{fig:motifcount}, with an average of 544.25 per motif and a standard deviation of 136.45. This set of motifs is based on two books which present a large number of visual motifs in cinema \cite{ballobergala2016motifs} and the public sphere \cite{salvado2023motifs}, and the ongoing research project FUCAV \cite{garin2021football} about the development of specific visual motifs in football during the Franco period. All three of these works place the concept of visual motif at the centre of their iconographic analysis. Inspired by this, our set of motifs was selected by the art experts on our team, according to how interesting they seem to analyse and the comparisons some of them generate among them (\eg \textit{Facing the Image}, \textit{Mirror}, and \textit{Woman in the Window}). The selected motifs raise many riveting questions, especially due to their frequent presence in contemporary visual culture. 

Each image is taken from a film, a TV show, a news report, a photograph, a painting, a comic, a sculpture, or any other form of visual medium. \Cref{fig:facingtheimage} presents some examples from the motif \textit{Facing the Image}. The proportions of each medium vary depending on the motif: some of them are more cinematographic and will have a large amount of examples from films, others are more journalistic or from the public sphere, and will contain more photographs.

\begin{figure}[tb]
  \centering
  \includegraphics[width=0.6\textwidth]{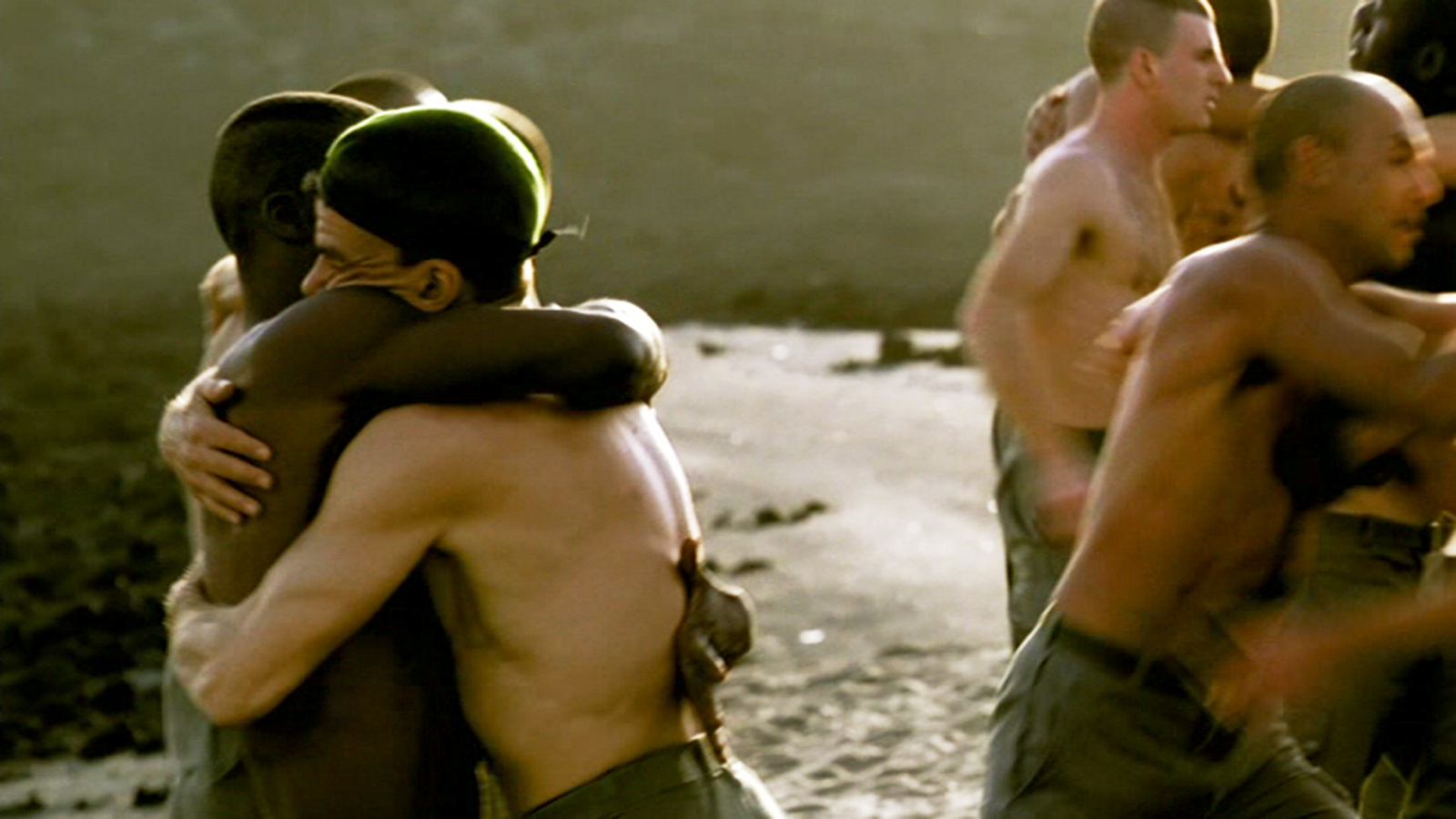}
  \caption{A frame from \emph{Beau Travail} (1999) by Claire Denis. Although the image clearly represents the \emph{Hug} motif, we can see that in the background, there is also a \emph{Brawl} going on, mostly off camera. The classification of this image within our dataset is therefore Primary Motifs: \emph{Hug}, and Secondary Motifs: \emph{Brawl}.}
  \label{fig:twomotifs}
\end{figure}

Originally, each image of the dataset corresponded to a single class, or motif, as is common in datasets used for classification. However, after internal discussions with the art experts on our team, we realised that some images can actually represent several motifs at once, sometimes even with different levels of importance. We needed these qualitative aspects of motif analysis to be represented quantitatively in the dataset, to be able to directly use this information in the training and evaluation of models. Each image therefore has a set of Primary Motifs and Secondary Motifs that it represents. An example image representing two motifs is given in \Cref{fig:twomotifs}. We consider the task as a multi-label classification, rather than a multi-class classification, that is to say that the output prediction for our model is a subset of the possible classes, instead of just one. By default, the correct predictions are the ones present in the Primary Motifs subset. In total, 159 images have Secondary Motifs, corresponding to 1.5\% of the dataset.

\begin{figure}
  \centering
  \setlength{\tabcolsep}{12pt}
  \begin{tabular}{cc}
  \includegraphics[height=4cm]{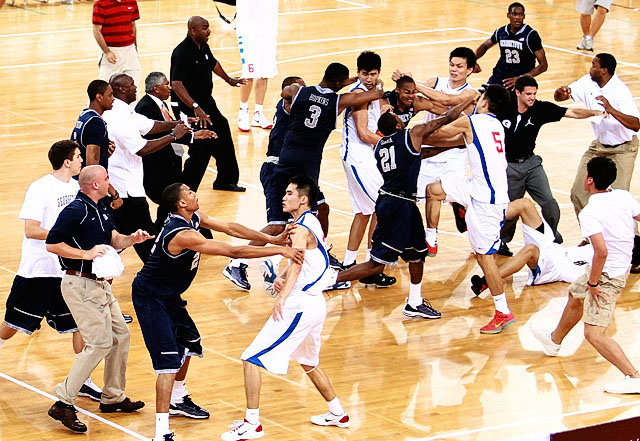} &
  \includegraphics[height=4cm]{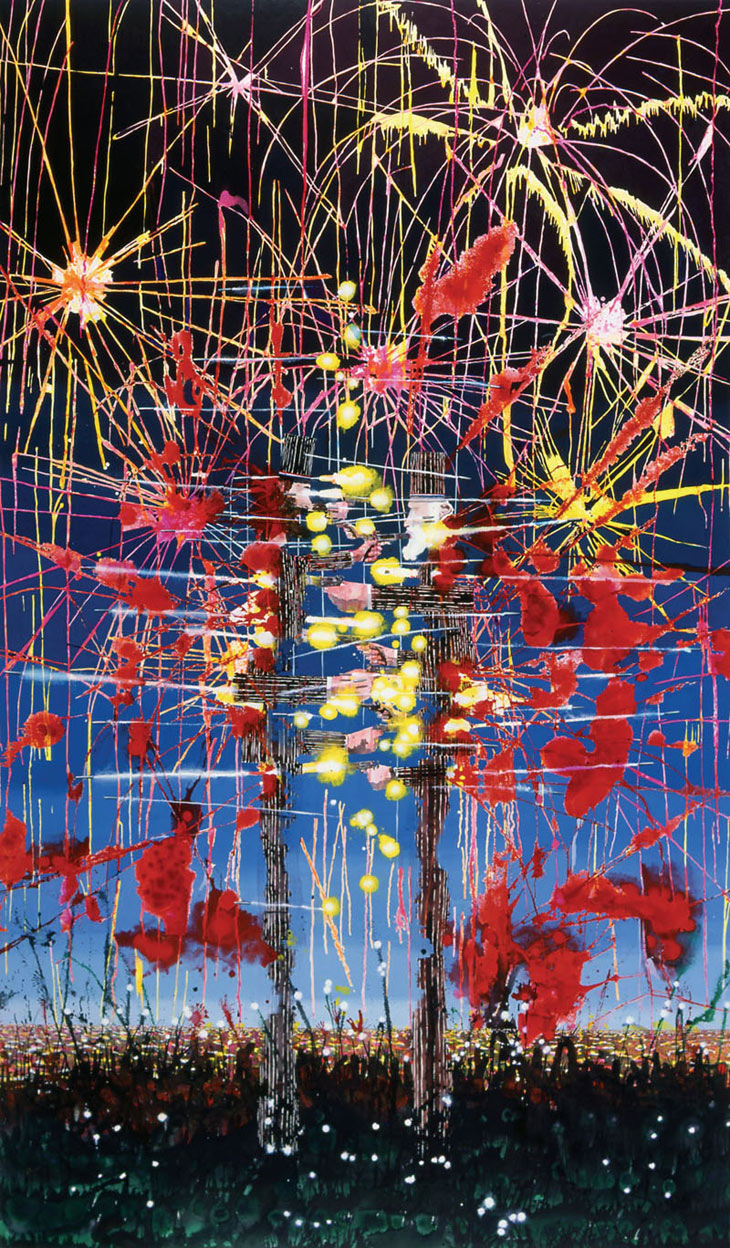}
  \end{tabular}
  \caption{Two examples from the \textit{Brawl} motif. On the left, a press photography from the BBC of a fight that broke out between the Georgetown Hoyas and the Bayi Rockets basketball teams in 2011. The typical sports setting, blatant pushing and grabbing of a large group of people, and wide shot showing the whole scene make this a Canonical image of the \textit{Brawl} motif. On the right, \emph{Duel (July 4th)} (2004) by Barnaby Furnas. The mix of colours, imagery of conflict, and general chaos present in this painting make it an instance of the \emph{Brawl} motif, but its abstract nature means that it is tagged as a Red Flag within the dataset.}
  \label{fig:rfandcan}
\end{figure}

Furthermore, within the set of images chosen to represent any given motif, some of them more obviously belong to the motif than others. That is to say, due to the intrinsic subjective nature of art analysis, a lack of context, or a very unusual composition, some images debatably represent the motif, or are right at the limit of what defines the motif. On the other hand, other images are clear, undeniable candidates as they follow the codes of the motif as forged throughout the history of art. This led us to tagging images with an indicator of how characteristic each one is of the motif(s) it is associated to in the dataset, allowing us to give more or less weight to these data samples during the training of a model. Images are therefore either tagged as Red Flag for the lesser representative ones, Canonical for the more typical ones, or not tagged for the standard ones (ultimately the most common tier of the three). An example of Red Flag and Canonical image for the motif \textit{Brawl} is presented in \Cref{fig:rfandcan}. In total, 1,159 images are considered Canonical, while 606 are considered Red Flag, corresponding to 10.8\% and 5.6\% of the dataset respectively.

This somewhat complex annotation system for our images allows for a more informative dataset, that better represents the subtlety and subjectivity of the intricate artistic concept that are visual motifs.

\section{Proposed Model}
\label{sec:model}

In the proposed method, from each image, we first extract initial CLIP-based features, which we project into a new embedding space. From this we obtain the probability the image has of belonging to each visual motif. The following subsections provide more details of our model.% each ingredient of our proposal is detailed.

\subsection{CLIP-Based Input Features}

Contrastive Language-Image Pretraining, CLIP \cite{radford2021CLIP}, and its open source counterpart, OpenCLIP \cite{cherti2023openclip}, are models that allow to relate images and text by encoding them in the same feature space, in such a way that cosine similarity of features directly correlates with semantic and visual similarities. For an input image, CLIP's output features therefore contain a lot of information on its contents, which we propose to leverage to determine the visual motifs present.

As a preliminary test, we divided the images of our dataset into 20 groups by clustering their normalised CLIP features with the $k$-means clustering algorithm. However, the resulting clusters in no way provide a classification of the images into motifs. Indeed, from what we have observed, CLIP seems to sort images superficially, according to aspects such as style (\eg black and white images, drawings, particular art styles) or presence of recognisable people (\eg Barack Obama). However, our interest lies in distinguishing images \emph{despite} these similarities, instead searching for iconographic motifs that can be found in images and visual data of any given style. Some motifs may be more partial to some styles than others, but it shouldn't be the deciding element for the classification. 

Despite searching for a classification that in some cases seems orthogonal to CLIP's initial clustering, we can extract the important information from these features and efficiently determine the motifs present in images by training our own classification head with an appropriate loss, representative of the subtleties that come with the analysis of visual motifs.

\subsection{Proposed Architecture} 

We use the EVA-CLIP \cite{sun2023evaclip} model EVA-01-CLIP-g/14+, a vision transformer composed of 40 attention layers each with 16 heads, available in OpenCLIP \cite{cherti2023openclip}, to generate the 1024 input features of our model. We freeze the CLIP model used throughout, that is to say that we don't perform any type of fine-tuning on the CLIP model, since we deem our dataset too small and the CLIP model too large for this to be effective. Instead, we train two successive linear layers, with a hidden layer of size 256 and a ReLU as the intermediate activation function, on the CLIP output features. The final layer contains 20 features, one for each of the 20 motifs. Our classification head therefore totals 267,540 parameters. Our model was trained using the Adam optimiser \cite{kingma2017adam} with no weight decay, over 200 epochs with learning rate 0.001 and batch size 256.

\subsection{Loss Function}
\label{sec:lossfunction}

Since our task is one of multi-label classification, our output is not a probability distribution over the classes, but rather a vector of probabilities, each representing how probable it is that the input belongs to the associated class. The prediction of our model will correspond to all motifs having an associated probability of at least 0.5. We adapt the Binary Cross Entropy loss as follows.

Let $N$ be the number of classes, that is, the number of visual motifs considered (in this paper, $N=20$). For any input image $I$, let $X^I = (x^I_1, ..., x^I_{N})\in \mathbb R^{N}$ be the associated output of the proposed model, returning one real number for each class. We directly relate each output feature $x^I_j$ to a probability of belonging to class $j$ via the sigmoid function:  $\sigma(x)= {e^x}/({1 + e^x})$, where $x \in \mathbb R$.

Using the dataset structure developed in \Cref{sec:dataset}, we define the target probabilities for $I$ as $T^I = (t^I_1, ..., t^I_{N})\in [0,1]^{N}$. For each class $j\in \{1,...,N\}$, let $t^I_j\in [0,1]$ be defined as:
\begin{equation}
t^I_j = \left\{
\begin{array}{ll}
    0 & \text{ if }j\text{ is not a correct motif for }I\\
    \mathit{SMT} & \text{ if }j\text{ is a Secondary Motif for }I\\
    1 & \text{ if }j\text{ is a Primary Motif for }I,
\end{array}
\right.
\end{equation}
where $\SMT$, Secondary Motif Target, is a hyperparameter, worth 0.5 by default.

For each class $j$, we can now compute the Binary Cross Entropy between the associated output and target probabilities:
\begin{equation}
\ell^I_j = -\left[t^I_j\log \sigma(x^I_j) + (1-t^I_j) \log(1-\sigma(x^I_j))\right].
\end{equation}
Note that this corresponds to the cross-entropy between the probability distributions given by $\{ t^I_j, 1 -  t^I_j\}$ and $\{ \sigma(x^I_j), 1 - \sigma(x^I_j)\}$.

Finally, we define for image $I$ a weight $w^I \in \{\mathit{RFW}, 1, \mathit{CW}\}$, according to how characteristic the image is. $\mathit{RFW}$ stands for Red Flag Weight and $\mathit{CW}$ stands for Canonical Weight.  
Suppose $I$ only has one Primary Motif $j'\in \{1,...,N\}$, then we define $w^I$ as:
\begin{equation}
w^I = \left\{
\begin{array}{ll}
    \mathit{RFW} & \text{ if }I\text{ is a Red Flag image for }j'\\
    1 & \text{ if }I\text{ is a standard image for }j'\\
    \mathit{CW} & \text{ if }I\text{ is a Canonical image for }j',
\end{array}
\right.
\end{equation}
with hyperparameters $\mathit{RFW} \leq 1$ (worth 0.5 by default), and $\mathit{CW} \geq 1$ (worth 2 by default). In the rare case that $I$ happens to have two or more Primary Motifs, we define $w^I$ as before, but with $j'$ the Primary Motif of $I$ with the highest representative tier.  

From this we compute the total Binary Cross Entropy loss associated to image $I$, as the mean Binary Cross Entropy over all classes, multiplied by the image's weight, that is:

\begin{equation}
\mathit{BCE}(X^I,T^I, w^I) = w^I\dfrac 1 N \sum\limits_{j = 1}^N \ell^I_j.
\end{equation}
Over a batch of images, the loss used is simply the mean of the $\mathit{BCE}$ loss applied to each element of the batch.

\section{Results and Analysis}

Due to the fact that our model outputs a set of predicted classes instead of just one, in this section, we first define some metrics for multi-label classification. Then, we analyse the performance of our model, and justify the structure of our dataset and the default values of hyperparameters through an ablation study. We also compare our results with some other baseline models.

\subsection{Evaluation Metrics}
\label{sec:evalmetrics}

Defining the accuracy of a multi-label classification model can be less straightforward than that of a typical multi-class classification, because even if the set of classes predicted isn't exactly correct, a portion of them may be. Based on \cite{luaces2012multilabelmetrics}, we consider a set of metrics to evaluate and compare models, using in different ways the set of ground truth and predicted motifs for each image.

We consider the example-based Precision and Recall metrics, along with the associated $F_1$-score, their harmonic mean. For an image $I$, let $O^I\subseteq \{1,...,N\}$ be the output set of predicted motifs, that is to say the motifs whose corresponding output probability given by the model applied to $I$ is at least 0.5. Let $\mathit{GT}^I\subseteq \{1,...,N\}$  be the set of ground truth motifs of $I$, which is, if unspecified, its set of Primary Motifs, but in some cases we consider both Primary and Secondary Motifs as correct. $\left|O^I \cap \mathit{GT}^I\right|$ is therefore the number of predictions that are correct, according to $I$'s ground truth motifs. Then, with $\mathit{TS}$ being the test set of images, a model's Precision $P$, Recall $R$, and $F_1$-score $F_1$ are defined as:
\begin{equation}
P = \dfrac{1}{|\mathit{TS}|} \sum\limits_ {I\in \mathit{TS}} \! \dfrac{|O^I \cap \mathit{GT}^I|}{|\mathit{GT}^I|},
\quad
R = \dfrac{1}{|\mathit{TS}|} \sum\limits_ {I\in \mathit{TS}} \! \dfrac{|O^I \cap \mathit{GT}^I|}{|O^I|},
\quad
F_1 = \dfrac{2  \! \cdot \!  P \! \cdot \!  R}{P+R}.
\end{equation}

We also consider our own metric that is closer to the classical accuracy: the Maximum Accuracy $\MA$, which represents the proportion of images for which the motif with the highest probability in the predictions is part of the ground truth set.

All of the metrics presented measure the performance of the model in the interval  $[0,1]$, with the higher values corresponding to better performing models. These metrics can also be analysed over Red Flag images and over Canonical images separately, to better understand how the model works on these subsets.

\begin{figure}
  \centering
  \begin{tabular}{rScScl}
  \begin{tabular}{r}PM: \textit{Mirror}\\Red Flag\\ \\ Predictions: \\ \textit{Mirror}: 0.9964 \\ \textit{Handshake}:\\ 0.5703\end{tabular} &
  \raisebox{-0.45\height}{\includegraphics[width=0.3\textwidth]{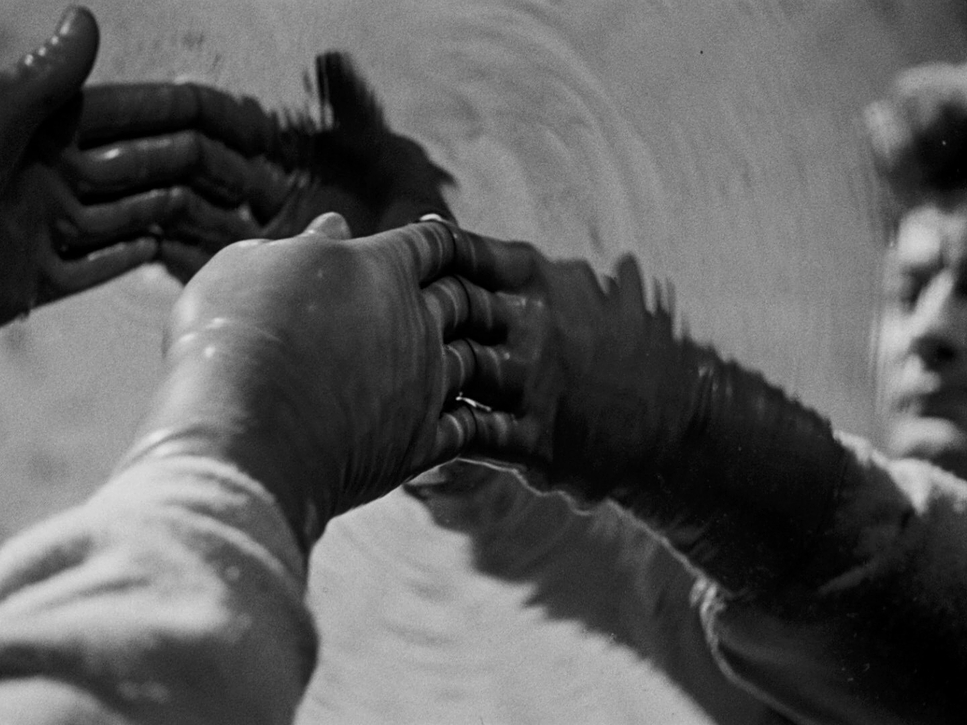}} &
  \raisebox{-0.45\height}{\includegraphics[width=0.3\textwidth]{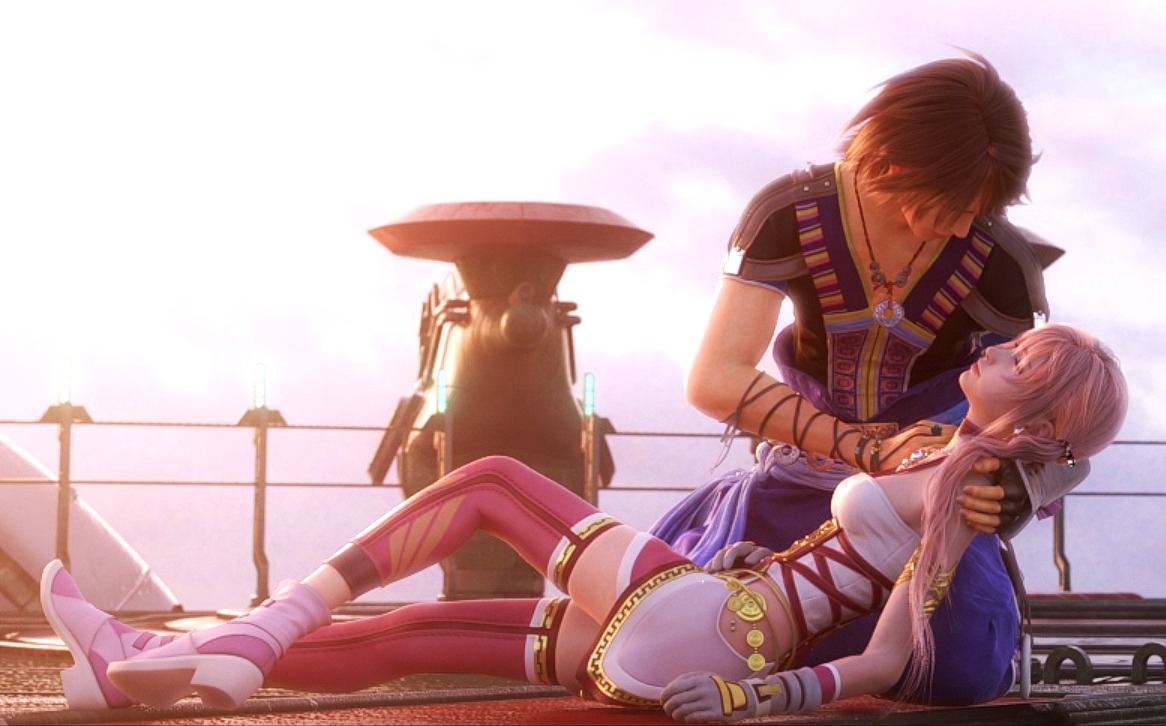}} &
  \begin{tabular}{l}PM: \textit{Pietà}\\Canonical\\ \\ Predictions: \\ \textit{Pietà}: 0.8936\\\textit{Hug}: 0.5341\end{tabular} \\
  \begin{tabular}{r}PM: \textit{Kiss}\\ \\ Predictions: \\ \textit{Hug}: 0.9793 \\ \textit{Kiss}: 0.7828\end{tabular} &
  \raisebox{-0.45\height}{\includegraphics[width=0.3\textwidth]{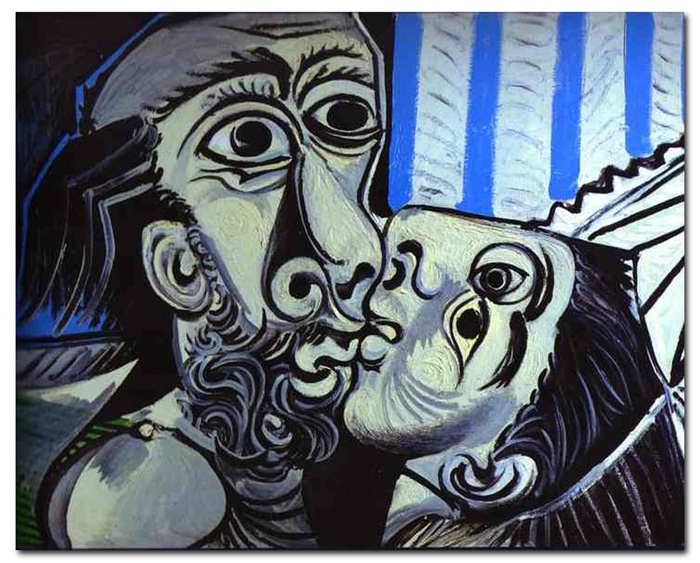}} &
  \raisebox{-0.45\height}{\includegraphics[width=0.3\textwidth]{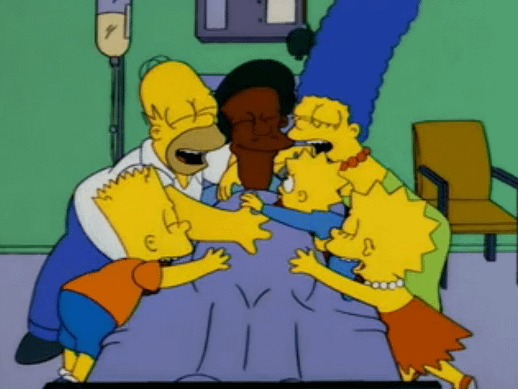}} &
  \begin{tabular}{l}PM: \textit{Hug}\\ \\Predictions: \\ \textit{Pietà}: 0.9772 \end{tabular}\\
  \end{tabular}
  \caption{Four examples of images from our test set, and their associated results after applying our model. For each image, we specify, when relevant, its Primary Motifs (PM), Secondary Motifs (SM), and tags (Red Flag or Canonical). We also give all motifs predicted by our model for that image, \ie those having a probability of at least 0.5 (\cf $O^I$ in \Cref{sec:evalmetrics}), and their probability. The images are, from top to bottom and left to right: a frame from \textit{Orpheus} (1950) by Jean Cocteau, a frame from the video game \emph{Final Fantasy VIII} (1999), the painting \textit{The Kiss} (1969) by Pablo Picasso, and a frame from S5E13 \textit{Homer and Apu} (1994) of \textit{The Simpsons}, by 	Mark Kirkland.}
  \label{fig:examples}
\end{figure}
\subsection{Results}

Our model, which we indicate with $(\star)$ in the rest of the paper, has an $F_1$ score of 0.9138 (0.9136 when counting Secondary Motifs as correct), with the Precision being 0.9055 and the Recall being 0.9223. Its Maximum Accuracy is 0.9459, meaning that almost 95\% of the most confident predictions among the images of the testing set were a Primary Motif for the corresponding image.

In most cases (88\% of the images in the test set), our model predicts the exact set of correct motifs. In \Cref{fig:examples}, we present some interesting predictions from our model on images of our test set: the first row shows two good predictions while the second row exhibits two failure cases. However, even in failure cases, certain similarities with the predicted motif can be inferred. For instance, a certain reminiscence of \textit{Hug} and \textit{Pietà} can be seen in the left and right images, respectively, of the second row. In general, we can see in some cases that incorrect predictions can make sense, and sometimes give insight into how the model works. For example, the secondary \textit{Handshake} prediction for the frame from \textit{Orpheus} (first row, left) is understandable due to the presence of two hands touching each other in that way. That confidence is low though, compared to that of the correct prediction \textit{Mirror} by our model.

\subsection{Ablation}

To justify the design choices in constructing the dataset and the loss function, we perform some ablation studies of the hyperparameters we introduced.

\subsubsection{Secondary Motif Target} 

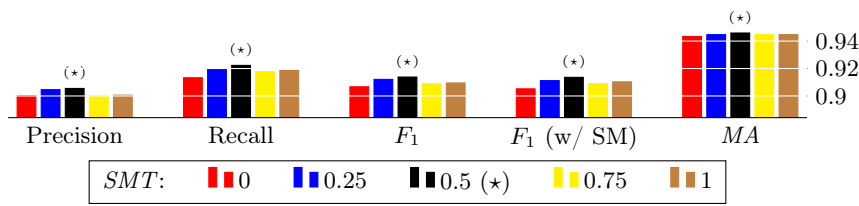
\begin{figure}
    \centering
    \pgfplotstabletranspose[input colnames to={SMT},colnames from={SMT}]{\transposeddata}{Metric_Graphs/SMT.dat}
\begin{tikzpicture}
  \centering
  \begin{axis}[
        ybar, axis on top,
        width = \textwidth,
        height = 3cm,
        bar width = 0.25cm,
        ymajorgrids, tick align=inside,
        major grid style={draw=white},
        enlarge y limits=0.35,
        axis x line*=bottom,
        axis y line*=right,
        y axis line style={opacity=0},
        tickwidth=0pt,
        enlarge x limits=true,
        legend style={
            at={(0.5,-.4)},
            anchor=north,
            legend columns=-1,
            /tikz/every even column/.append style={column sep=0.5cm},
        },
        legend cell align={left},
        cycle list name = color list,
        no markers,
        xtick={0, 0.25, 0.5, 0.75, 1},
        ytick = {0.9,0.92,0.94},
        xticklabels={Precision, Recall, $F_1$, $F_1$ (w/ SM), $\MA$},
        xmin = 0 , xmax = 1,
    ]
    \addlegendimage{empty legend}
    \addplot+[fill] table[x expr = \lineno/4, y = 0] {\transposeddata};
    \addplot+[fill] table[x expr = \lineno/4, y = 0.25] {\transposeddata};
    \addplot+[fill] table[x expr = \lineno/4, y = 0.5] {\transposeddata}
    node [pos = 0, above, color = black] {\tiny $(\star)$}
    node [pos = 0.25, above, color = black] {\tiny $(\star)$}
    node [pos = 0.5, above, color = black] {\tiny $(\star)$}
    node [pos = 0.75, above, color = black] {\tiny $(\star)$}
    node [pos = 1, above, color = black] {\tiny $(\star)$};
    \addplot+[fill] table[x expr = \lineno/4, y = 0.75] {\transposeddata};
    \addplot+[fill] table[x expr = \lineno/4, y = 1] {\transposeddata};
    \legend{$\SMT$:, 0, 0.25, 0.5 ($\star$), 0.75, 1}    
    
  \end{axis}
  \end{tikzpicture}
    \caption{Performance of models trained with various values of the Secondary Motif Target hyperparameter, $\SMT$. We analyse the Precision, Recall, $F_1$-score, $F_1$-score counting Secondary Motifs as correct, and Maximum Accuracy of each model. Our model is indicated with $(\star)$.}
    \label{fig:SMTablation}
\end{figure}

In \Cref{fig:SMTablation}, we compare the performance of models when varying the hyperparameter $\mathit{SMT}$ (see \cref{sec:lossfunction}), \ie the target probability given to Secondary Motifs during training. All other hyperparameters are left unchanged.

The value $\SMT = 0.5$ gives the best results across all considered metrics, showing that giving an intermediate probability target to Secondary Motifs improves the results of the model, whether we consider Secondary Motifs as correct or not in the evaluation stage. Note that for $\SMT = 1$ in \Cref{fig:SMTablation}, Secondary Motifs and Primary Motifs are considered as equals in the training stage, but our proposed model outperforms it even when Secondary Motifs are considered correct when evaluating ($F_1$ w/ SM). Similarly, when Secondary Motifs are essentially ignored during training when $\SMT = 0$, results are also inferior to our proposed model. Considering Secondary Motifs as neither correct nor incorrect, but rather as somewhere in between, is therefore the best approach.

\subsubsection{Red Flag and Canonical Weights}

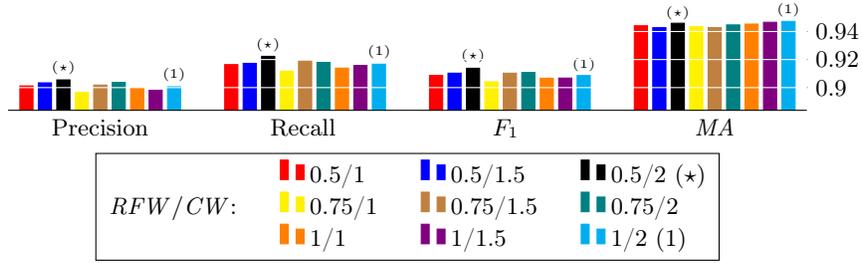
\begin{figure}
  \centering
  \pgfplotstabletranspose[input colnames to={RFW-CW},colnames from={RFW-CW}]{\transposeddata}{Metric_Graphs/RFW-CW.dat}
\begin{tikzpicture}
  \centering
  \begin{axis}[
        ybar, axis on top,
        width = \textwidth,
        height = 3cm,
        bar width = 0.175cm,
        ymajorgrids, tick align=inside,
        major grid style={draw=white},
        enlarge y limits=0.25,
        axis x line*=bottom,
        axis y line*=right,
        y axis line style={opacity=0},
        tickwidth=0pt,
        enlarge x limits=0.15,
        legend style={
            at={(0.5,-0.4)},
            anchor=north,
            legend columns=4,
            /tikz/every even column/.append style={column sep=0.5cm},
        },
        legend cell align={left},
        cycle list name = color list,
        no markers,
        xtick={0, 0.33, 0.66, 1},
        ytick={0.88, 0.9, 0.92, 0.94},
        xticklabels={Precision, Recall, $F_1$, $\MA$},
        xmin = 0 , xmax = 1,
    ]
    \addlegendimage{empty legend}
    \addplot+[fill] table[x expr = \lineno/3, y = 0.5-1] {\transposeddata};
    \addplot+[fill] table[x expr = \lineno/3, y = 0.5-1.5] {\transposeddata};
    \addplot+[fill] table [x expr = \lineno/3, y = 0.5-2] {\transposeddata}
    node[color = black] at (-6,180) {\tiny $(\star)$}
    node[color = black] at (27.3,340) {\tiny $(\star)$}
    node[color = black] at (60.7, 260) {\tiny $(\star)$}
    node[color = black] at (94, 570){\tiny $(\star)$};
    \addlegendimage{empty legend}
    \addplot+[fill] table[x expr = \lineno/3, y = 0.75-1] {\transposeddata};
    \addplot+[fill] table[x expr = \lineno/3, y = 0.75-1.5] {\transposeddata};
    \addplot+[fill] table[x expr = \lineno/3, y = 0.75-2] {\transposeddata};
    \addlegendimage{empty legend}
    \addplot+[fill] table[x expr = \lineno/3, y = 1-1] {\transposeddata};
    \addplot+[fill] table[x expr = \lineno/3, y = 1-1.5] {\transposeddata};
    \addplot+[fill] table[x expr = \lineno/3, y = 1-2] {\transposeddata}
    node[color = black] at (12,130) {\tiny $(1)$}
    node[color = black] at (45.4,290) {\tiny $(1)$}
    node[color = black] at (78.7, 200) {\tiny $(1)$}
    node[color = black] at (112, 590){\tiny $(1)$};
    \legend{\,, 0.5/1, 0.5/1.5, 0.5/2 ($\star$), $\RFW/\CW$:, 0.75/1, 0.75/1.5, 0.75/2, \,, 1/1, 1/1.5, 1/2 (1)}

  \end{axis}
  \end{tikzpicture}
  \caption{Performance of models trained with various values of the hyperparameters Red Flag Weight, $\RFW$, and Canonical Weight, $\CW$. We analyse the Precision, Recall, $F_1$-score, and Maximum Accuracy of each model. Model (1) is compared with our model, $(\star)$, in \Cref{tab:detailedperformance}.}
  \label{fig:RFWandCWablation}
\end{figure}

In \Cref{fig:RFWandCWablation}, we compare the performance of models when modifying the hyperparameters $\mathit{RFW}$ and $\CW$ (see \cref{sec:lossfunction}), \ie the weights given to images according to how representative of their Primary Motifs they are. All other hyperparameters are left unchanged.

We see that our choices for the defaults of $\mathit{RFW}$ and $\mathit{CW}$, 0.5 and 2 respectively, generally give the best results. In this case, they outperform all over values tested for the first three metrics, and come third of nine for the Maximum Accuracy. Note the ranking of the model trained with $\mathit{RFW} = 1$ and $\mathit{CW} = 1$: 7th, 8th and 8th of 9 for Precision, Recall, and $F_1$-score respectively (and 4th for Maximum Accuracy). Therefore, using a classification of how characteristic each image is of its Primary Motifs allows to improve results.

\subsubsection{Architecture} 
\begin{figure}
  \centering
  \pgfplotstabletranspose[input colnames to={HL},colnames from={HL}]{\transposeddata}{Metric_Graphs/CLIP2.dat}
\begin{tikzpicture}
  \centering
  \begin{axis}[
        ybar, axis on top,
        width = \textwidth,
        height = 3cm,
        bar width = 0.23cm,
        ymajorgrids, tick align=inside,
        major grid style={draw=white},
        enlarge y limits=0.35,
        axis x line*=bottom,
        axis y line*=right,
        y axis line style={opacity=0},
        tickwidth=0pt,
        enlarge x limits=0.15,
        legend style={
            at={(0.5,-0.4)},
            anchor=north,
            legend columns=-1,
            /tikz/every even column/.append style={column sep=0.5cm},
        },
        legend cell align={left},
        cycle list name = color list,
        no markers,
        xtick={0, 0.33, 0.66, 1},
        ytick={0.88, 0.9, 0.92, 0.94},
        xticklabels={Precision, Recall, $F_1$, $\MA$},
        xmin = 0 , xmax = 1,
    ]
    \addlegendimage{empty legend}
    \addplot+[fill] table[x expr = \lineno/3, y = 64] {\transposeddata}
    node[color = black] at (-9.1,140) {\tiny $(2)$}
    node[color = black] at (24.2,290) {\tiny $(2)$}
    node[color = black] at (57.5, 220) {\tiny $(2)$}
    node[color = black] at (90.9, 650){\tiny $(2)$};
    \addplot+[fill] table[x expr = \lineno/3, y = 128] {\transposeddata};
    \addplot+[fill] table[x expr = \lineno/3, y = 256] {\transposeddata}
    node[color = black] at (-1.8,230) {\tiny $(\star)$}
    node[color = black] at (31.5,380) {\tiny $(\star)$}
    node[color = black] at (64.9, 300) {\tiny $(\star)$}
    node[color = black] at (98.2, 610){\tiny $(\star)$};
    \addplot+[fill] table[x expr = \lineno/3, y = 512] {\transposeddata};
    \addplot+[fill] table[x expr = \lineno/3, y = 768] {\transposeddata};
    \addplot+[fill] table[x expr = \lineno/3, y = 1024] {\transposeddata};
    \legend{$\mathit{HL}$:, 64 (2), 128, 256 $(\star)$, 512, 768, 1024}

  \end{axis}
  \end{tikzpicture}
  \caption{Performance of models trained with CLIP features having two layers, with varying size of the hidden layer ($\mathit{HL}$). We report the Precision, Recall, $F_1$-score, and Maximum Accuracy. Model (2) is compared with our model, $(\star)$, in \Cref{tab:detailedperformance}.}
  \label{fig:clip2layersablation}
\end{figure}
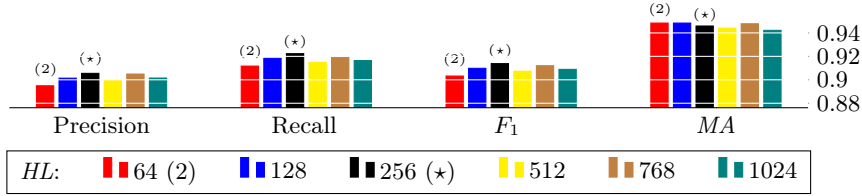

We compare models trained with networks of different sizes. In \Cref{fig:clip2layersablation} we show results of models having 2 layers as before, but with varying size of the hidden layer. We show that a size of 256 gives the best Precision, Recall, and $F_1$ of all the sizes tested.

\begin{figure}
  \centering
  \pgfplotstabletranspose[input colnames to={HLs},colnames from={HLs}]{\transposeddata}{Metric_Graphs/CLIP-3-4.dat}
\begin{tikzpicture}
  \centering
  \begin{axis}[
        ybar, axis on top,
        width = \textwidth,
        height = 3cm,
        bar width = 0.15cm,
        ymajorgrids, tick align=inside,
        major grid style={draw=white},
        enlarge y limits=0.2,
        axis x line*=bottom,
        axis y line*=right,
        y axis line style={opacity=0},
        tickwidth=0pt,
        enlarge x limits=0.15,
        legend style={
            name = leg1, draw = none,
            at={(0.5,-0.5)},
            anchor=north,
            legend columns=4,
            /tikz/every even column/.append style={column sep=0.5cm},
        },
        legend cell align={left},
        cycle list name = color list,
        no markers,
        xtick={0, 0.33, 0.66, 1},
        ytick={0.88, 0.91, 0.94},
        xticklabels={Precision, Recall, $F_1$, $\MA$},
        xmin = 0 , xmax = 1,
    ]
    \addplot+[fill] table[x expr = \lineno/3, y = 256-64] {\transposeddata};
    \addplot+[fill] table[x expr = \lineno/3, y = 256-128] {\transposeddata};
    \addplot+[fill] table[x expr = \lineno/3, y = 256-256] {\transposeddata};
    \addplot+[fill] table[x expr = \lineno/3, y = 512-128] {\transposeddata};
    \addlegendimage{empty legend}
    \addplot+[fill] table[x expr = \lineno/3, y = 768-128] {\transposeddata};
    \addplot+[fill] table[x expr = \lineno/3, y = 768-256] {\transposeddata}
    node[color = black] at (1.4,340) {\tiny $(3)$}
    node[color = black] at (34.7,480) {\tiny $(3)$}
    node[color = black] at (68, 400) {\tiny $(3)$}
    node[color = black] at (101.3, 630){\tiny $(3)$};
    \addlegendimage{empty legend}
    \addplot+[fill] table[x expr = \lineno/3, y = 512-128-64] {\transposeddata};
    \addplot+[fill] table[x expr = \lineno/3, y = 768-128-128] {\transposeddata};
    \addplot+[fill] table[x expr = \lineno/3, y = 768-128-256] {\transposeddata};
    \addplot+[fill] table[x expr = \lineno/3, y = 768-256-256] {\transposeddata};
    \legend{$256/64/-$, $256/128/-$, $256/256/-$, $512/128/-$, \,, $768/128/-$, $768/256/- (3)$, \,, $512/128/64$, $768/128/128$, $768/128/256$, $768/256/256$}
  \end{axis}
  \begin{axis}[
      hide axis,
      legend columns=-1,
      legend style={
        name=leg2,draw=none,
        anchor=north,
        at={(0.75,-.37)}
        }         
      ]
    \addplot[forget plot] coordinates {(0,0)};

    \addlegendimage{empty legend};
    \addlegendentry{$\mathit{HL_1}/\mathit{HL_2}/\mathit{HL_3}$}
    \end{axis}

    \node [fit=(leg1)(leg2),draw] {};
  \end{tikzpicture}
  \caption{Performance of the best models trained with CLIP features having three or four layers, with varying size of the hidden layers ($\mathit{HL_1}$, $\mathit{HL_2}$, and $\mathit{HL_3}$). We report the Precision, Recall, $F_1$-score, and Maximum Accuracy. Model (3) is compared with our model in \Cref{tab:detailedperformance}.}
  \label{fig:clip3and4layersablation}
\end{figure}
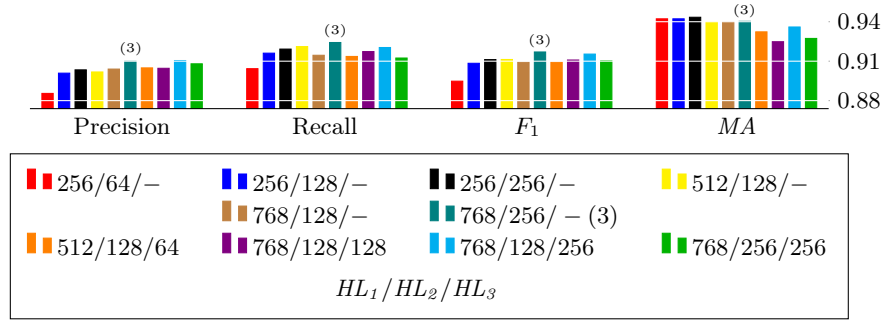

We also trained models with 3 and 4 layers, and present the models with the highest performances for each of the four metrics, for the corresponding number of layers, in \Cref{fig:clip3and4layersablation}. The three-layer model highlighted as (3) in the table has slightly better Precision, Recall, and $F_1$-score than our model, but it also has a lower Maximum Accuracy.This model is therefore comparable when evaluated on the whole testing set, but we can see for example in \Cref{tab:detailedperformance} that our model outperforms it on Red Flag images, a more difficult subset of images to classify.

\subsection{Baseline Models}

\subsubsection{DINOv2} 
\begin{figure}
  \centering
  \pgfplotstabletranspose[input colnames to={HLs},colnames from={HLs}]{\transposeddata}{Metric_Graphs/Dino.dat}
\begin{tikzpicture}
  \centering
  \begin{axis}[
        ybar, axis on top,
        width = \textwidth,
        height = 2.5cm,
        bar width = 0.25cm,
        ymajorgrids, tick align=inside,
        major grid style={draw=white},
        enlarge y limits=0.75,
        axis x line*=bottom,
        axis y line*=right,
        y axis line style={opacity=0},
        tickwidth=0pt,
        enlarge x limits=0.15,
        legend style={
            at={(0.5,-0.6)},
            anchor=north,
            legend columns=3,
            /tikz/every even column/.append style={column sep=0.5cm},
        },
        legend cell align={left},
        cycle list name = color list,
        no markers,
        xtick={0, 0.33, 0.66, 1},
        ytick={0.86, 0.9},
        xticklabels={Precision, Recall, $F_1$, $\MA$},
        xmin = 0 , xmax = 1,
    ]
    \addlegendimage{empty legend}
    \addplot+[fill] table[x expr = \lineno/3, y = 512] {\transposeddata}
    node[color = black] at (-7.9,285) {\tiny $(4)$}
    node[color = black] at (25.5,370) {\tiny $(4)$}
    node[color = black] at (58.8, 325) {\tiny $(4)$}
    node[color = black] at (92.2, 825){\tiny $(4)$};
    \addplot+[fill] table[x expr = \lineno/3, y = 1024] {\transposeddata};
    \addlegendimage{empty legend}
    \addplot+[fill] table[x expr = \lineno/3, y = 512-256] {\transposeddata}
    node [pos = 0, above, color = black] {\tiny $(5)$}
    node [pos = 0.25, above, color = black] {\tiny $(5)$}
    node [pos = 0.5, above, color = black] {\tiny $(5)$}
    node [pos = 0.75, above, color = black] {\tiny $(5)$}
    node [pos = 1, above, color = black] {\tiny $(5)$};
    \addlegendimage{empty legend}
    \addlegendimage{empty legend}
    \addplot+[fill] table[x expr = \lineno/3, y = 256-128-128] {\transposeddata}
    node[color = black] at (4,590) {\tiny $(6)$}
    node[color = black] at (37.3,630) {\tiny $(6)$}
    node[color = black] at (70.7, 610) {\tiny $(6)$}
    node[color = black] at (104.1, 750){\tiny $(6)$};
    \addplot+[fill] table[x expr = \lineno/3, y = 512-128-256] {\transposeddata};
    \legend{\,, $512/-/- (4)$, $1024/-/-$, $\mathit{HL_1}/\mathit{HL_2}/\mathit{HL_3}$, $512/256/- (5)$, \,, \,, $256/128/128 ~(6)$, $512/128/256$}
  \end{axis}
  \end{tikzpicture}
  \caption{Performance of the best models trained with DINOv2 features having two, three, or four layers, with varying size of the hidden layers ($\mathit{HL_1}$, $\mathit{HL_2}$, and $\mathit{HL_3}$). We report the Precision, Recall, $F_1$-score, and Maximum Accuracy of each model. Models (4), (5), and (6) are compared with our model in \Cref{tab:detailedperformance}.}
  \label{fig:dinoablation}
\end{figure}
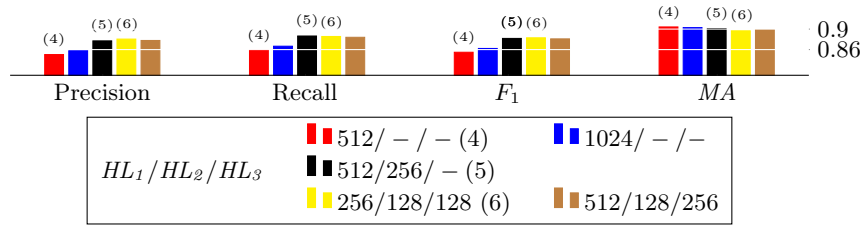

To be able to compare the performance of our model and the efficiency of CLIP features, we train models with the same architectures and hyperparameters, but using other features as input: those of DINOv2 \cite{oquab2024dinov2}, which yields a vector of 1536 features, compared to EVA-CLIP's 1024 in our model. We summarise our findings in \Cref{fig:dinoablation} by presenting the models that have the highest performance for one of the four metrics, for the corresponding number of layers. We note good results, but none of the models surpass our model trained on CLIP features. In fact, none of them reach 0.89 $F_1$-score, nor 0.91 Maximum Accuracy.

\subsubsection{Detectron2} As another comparison, we tried training models using features extracted from a well-known object detection library, Detectron2 \cite{wu2019detectron2}. In particular, we use the Mask R-CNN model \cite{he2017mask} with a ResNet-50-FPN backbone \cite{lin2017fpn}, trained on the COCO dataset \cite{lin2015coco}, where each layer of the Feature Pyramid Network encodes objects of different scales. The smallest layer yields 256 channels, each containing a $13\times 20$ matrix of features, giving us 66560 features in total, \ie 65 times more than the CLIP features used. Since these features are obtained via convolutional networks, they are local (encode a region of the input image), and convolutional networks can be applied to them to classify them.

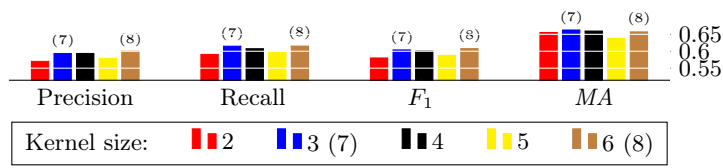
\begin{figure}
  \centering
  \pgfplotstabletranspose[input colnames to={Kernel},colnames from={Kernel}]{\transposeddata}{Metric_Graphs/Detectron2.dat}
\begin{tikzpicture}
  \centering
  \begin{axis}[
        ybar, axis on top,
        width = 0.85\textwidth,
        height = 2.5cm,
        bar width = 0.23cm,
        ymajorgrids, tick align=inside,
        major grid style={draw=white},
        enlarge y limits=0.6,
        axis x line*=bottom,
        axis y line*=right,
        y axis line style={opacity=0},
        tickwidth=0pt,
        enlarge x limits=0.15,
        legend style={
            at={(0.5,-0.6)},
            anchor=north,
            legend columns=-1,
            /tikz/every even column/.append style={column sep=0.5cm},
        },
        legend cell align={left},
        cycle list name = color list,
        no markers,
        xtick={0, 0.33, 0.66, 1},
        ytick ={0.65, 0.55, 0.6},
        xticklabels={Precision, Recall, $F_1$, $\MA$},
        xmin = 0 , xmax = 1,
    ]
    \addlegendimage{empty legend}
    \addplot+[fill] table[x expr = \lineno/3, y = 2] {\transposeddata};
    \addplot+[fill] table[x expr = \lineno/3, y = 3] {\transposeddata}
    node[color = black] at (-4.4,605) {\tiny $(7)$}
    node[color = black] at (28.9,810) {\tiny $(7)$}
    node[color = black] at (62.3, 705) {\tiny $(7)$}
    node[color = black] at (95.6, 1285){\tiny $(7)$};
    \addplot+[fill] table[x expr = \lineno/3, y = 4] {\transposeddata};
    \addplot+[fill] table[x expr = \lineno/3, y = 5] {\transposeddata};
    \addplot+[fill] table[x expr = \lineno/3, y = 6] {\transposeddata}
    node[color = black] at (8.9, 650) {\tiny $(8)$}
    node[color = black] at (42.2, 840) {\tiny $(8)$}
    node[color = black] at (75.6, 765) {\tiny $(8)$}
    node[color = black] at (108.9, 1255){\tiny $(8)$};
    \legend{Kernel size:, 2, 3 (7), 4, 5, 6 (8)}

  \end{axis}
  \end{tikzpicture}
  \caption{Performance of models trained with Detectron2 (Mask R-CNN model with a ResNet-50-FPN backbone) features, using two convolutional layers with varying kernel size and two linear layers. We report the Precision, Recall, $F_1$-score, and Maximum Accuracy of each model. Models (7) and (8) are compared with our model in \Cref{tab:detailedperformance}.}
  \label{fig:detectron2ablation}
\end{figure}

Using the same loss and evaluation methods, we were not able to train any models with results even comparable to the previously presented ones. Using two convolutional layers followed by two linear layers gives the best results. We vary the kernel size used, as shown in \Cref{fig:detectron2ablation}, and we highlight models (7) and (8) which give the best scores for the metrics considered. It seems that exclusively analysing the objects of the scene is not sufficient to accurately determine visual motifs. Including information on how these objects relate to each other within the frame, as in \cite{krishna2017visualgenome}, could be an interesting route to improving results.

\begin{table}%[tb]
  \caption{Detailed performance of our model, $(\star)$, and select models trained with CLIP, DINOv2, and Detectron2 features. We report the $F_1$-score and Maximum Accuracy of each model, over the entire test set, the Red Flag images of the test set, and the Canonical images of the test set.}
  \label{tab:detailedperformance}
  \centering
  \begin{tabular}{@{}c@{\hskip 0.4cm}c@{\hskip 0.4cm}ccc@{\hskip 0.4cm}ccc@{\hskip 0.4cm}c@{}}
    \toprule
     & \multicolumn{2}{c}{Test set} & \hspace{0.3cm} & \multicolumn{2}{c}{Red Flag images} & \hspace{0.3cm} & \multicolumn{2}{c}{Canonical images} \\[\aboverulesep] 
    \cline{2-3}  \cline{5-6} \cline{8-9}\\[-0.3cm]
    Model & $F_1$ & $\MA$ & & $F_1$ & $\MA$ & & $F_1$ & $\MA$ \\[\aboverulesep]
    \cline{2-3}  \cline{5-6} \cline{8-9}\\[-0.3cm]
    CLIP $(\star)$ & 0.9138 & 0.9459 & & 0.8338 & \textbf{0.9213} & & 0.9051 & 0.9483\\
    CLIP (1) & 0.9086 & 0.9472 & & 0.8367 & 0.9101 & & 0.8979 & \textbf{0.9540}\\
    CLIP (2) & 0.9032 & \textbf{0.9484} & & 0.8133 & \textbf{0.9213} & & 0.8965 & 0.9425\\
    CLIP (3) & \textbf{0.9170} & 0.9403 & & 0.8103 & 0.8764 & & \textbf{0.9210} & 0.9368\\
    DINOv2 (4) & 0.8548 & 0.9042 & & 0.7755 & 0.8750 & & 0.8742 & 0.9064\\
    DINOv2 (5) & 0.8818 & 0.9010 & & 0.7953 & 0.8977 & & 0.8860 & 0.8947 \\
    DINOv2 (6) & 0.8830 & 0.8966 & & \textbf{0.8523} & 0.8750 & & 0.8801 & 0.8889\\
    Detectron2 (7) & 0.6039 & 0.6627 & & 0.4042 & 0.4494 & & 0.6996 & 0.7414\\
    Detectron2 (8) & 0.6077 & 0.6571 & & 0.3846 & 0.4270 & & 0.6765 & 0.7356\\
  \bottomrule
  \end{tabular}
\end{table}
\Cref{tab:detailedperformance} compares our model $(\star)$ with other models trained with CLIP, and the best baseline models, all indicated with numbers in the corresponding figures. We compare the performance of these models not only on the test set, but also over the test set's Red Flag and Canonical images, separately. We can conclude that, compared to Detectron2, CLIP and DINOv2 provide richer features that are more adapted to our purposes. Overall, the best results are obtained with CLIP features, and our model ranks in the top 3 models of 9, for the 6 metrics analysed. For Red Flag cases in particular, which are, according to experts, more difficult to identify, our model gives the highest probability to a correct motif more often than nearly any other model, by a considerable margin. Its performance on Canonical images is also very good for both the metrics considered.

\section{Conclusions and Future Work}

In this paper, we have proposed an approach for recognising visual motifs that appear in various media, shaping a specific visual culture. To that goal, we have developed an image dataset, from which we extracted corresponding features using pre-existing models to train multi-label classification networks. This being a collaborative work with experts in art history and the aesthetic of cinema, the development of the dataset and the qualitative evaluation of models go hand in hand with the loss function, the evaluation metrics, and the structure of the dataset used, thus ensuring they all serve the desired behaviour of the model. 

We showed that by applying a CLIP model to the images of the dataset, the extracted features could be used to design and train an efficient multi-label classification model, able to recognise the pre-selected motifs with an $F_1$-score of over 0.91. In fact, our model does so with a surprisingly small classification head, containing only 2 linear layers, showing the potential of CLIP features for the understanding of aesthetic and artistic high-level concepts. In comparison, models trained on features extracted from DINOv2 or Detectron2 get significantly worse evaluations, despite being encoded in higher-dimensional spaces

This being a project in its first stages, there are several ways in which the model may be expanded or improved. First of all, the dataset can be extended, both in number of motifs and number of images per motif. It would be interesting to see how the performance of the model varies when new motifs are added. That being said, the current performance using CLIP is very satisfactory, and the meaning behind the features of our dataset can be dug into further: what part of the images does our model focus on? Is the composition of the frame taken into account and is it represented in our features? From this, can we determine composition prototypes for each of the motifs? Or are more hierarchical or graph-based architectures necessary for such a task? Finally, apart from the quantitative measures presented, we plan to assess our findings by conducting a user study, asking experts in art, narrative, cinema, and comparative media to qualitatively evaluate the performance of our model against new images, coming from different types of media, and from film and art history.

\section*{Acknowledgements}

The authors acknowledge support by MICINN/FEDER UE project, ref. PID2021-127643NB-I00, and by Maria de Maeztu Units of Excellence Programme CEX2021-001195-M, funded by MICIU/AEI /10.13039/501100011033. They also thank the reviewers for their valuable comments and suggestions that helped to improve the manuscript.


\bibliographystyle{splncs04}
\bibliography{main}
\end{document}